\pdfoutput=1
\documentclass[11pt]{article}

\usepackage{acl}

\usepackage{times}
\usepackage{latexsym}

\usepackage[T1]{fontenc}
\usepackage[utf8]{inputenc}

\usepackage{microtype}

\usepackage{graphicx}
\usepackage{subcaption}
\usepackage{amssymb}
\usepackage{booktabs}

\title{DistilQwen2.5: Industrial Practices of Training Distilled Open Lightweight Language Models}

\author{Chengyu Wang\thanks{\ \ \ C. Wang and J. Yan contributed equally to this work.}, Junbing Yan\footnotemark[1], Yuanhao Yue, Jun Huang\\
  Alibaba Cloud Computing\\
  \texttt{chengyu.wcy@alibaba-inc.com}\\}

\begin{document}
\maketitle
\begin{abstract}
Enhancing computational efficiency and reducing deployment costs for large language models (LLMs) have become critical challenges in various resource-constrained scenarios. In this work, we present \emph{DistilQwen2.5}, a family of distilled, lightweight LLMs derived from the public \emph{Qwen2.5} models. These distilled models exhibit enhanced instruction-following capabilities compared to the original models based on a series of distillation techniques that incorporate knowledge from much larger LLMs. In our industrial practice, we first leverage powerful proprietary LLMs with varying capacities as multi-agent teachers to select, rewrite, and refine instruction-response pairs that are more suitable for student LLMs to learn. After standard fine-tuning, we further leverage a computationally efficient model fusion approach that enables student models to progressively integrate fine-grained hidden knowledge from their teachers. Experimental evaluations demonstrate that the distilled models possess significantly stronger capabilities than their original checkpoints. Additionally, we present use cases to illustrate the applications of our framework in real-world scenarios. To facilitate practical use, we have released all the \emph{DistilQwen2.5} models to the open-source community.
\footnote{Our trained lightweight models and our processed large instruction-following dataset are released in HuggingFace.}
\end{abstract}

\section{Introduction}

Large language models (LLMs) have emerged as a transformative technology in NLP, powering a wide array of applications from machine translation to conversational agents~\cite{DBLP:journals/corr/abs-2303-18223}. However, the rise of LLMs has been accompanied by several challenges, notably the substantial computational resource requirements and high deployment costs. Reducing the parameter sizes of LLMs while maintaining or even improving performance has become a critical area of research.

Knowledge distillation (KD) is a promising approach to addressing these challenges by transferring knowledge from a larger model (the teacher) to a smaller model (the student)~\cite{DBLP:journals/corr/abs-2402-13116}. Previous works have primarily focused on specific KD techniques to develop more robust student models~\cite{DBLP:conf/acl/HsiehLYNFRKLP23,DBLP:conf/iclr/Gu0WH24,DBLP:conf/emnlp/YueWHW24,DBLP:conf/emnlp/ZhangZS0X24}. However, there is a lack of studies investigating good industrial practices that create a series of distilled lightweight LLMs with varying sizes and capacities.

\begin{figure}
    \centering
    \begin{subfigure}{0.23\textwidth}
        \centering
        \includegraphics[width=\textwidth]{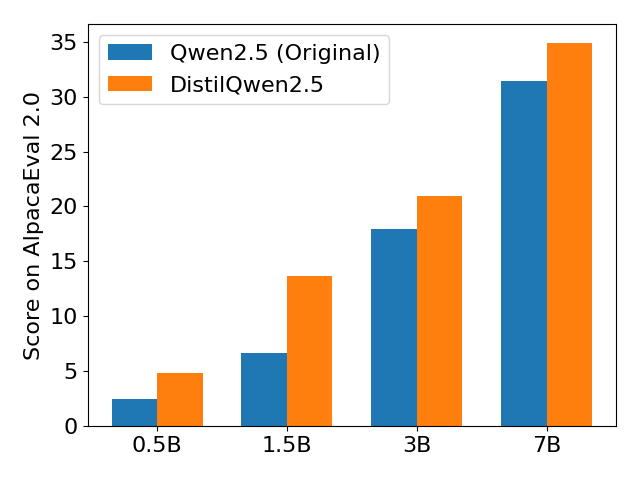}
        %\caption{AlpacaEval 2.0}
    \end{subfigure}
    \hfill
    \begin{subfigure}{0.23\textwidth}
        \centering
        \includegraphics[width=\textwidth]{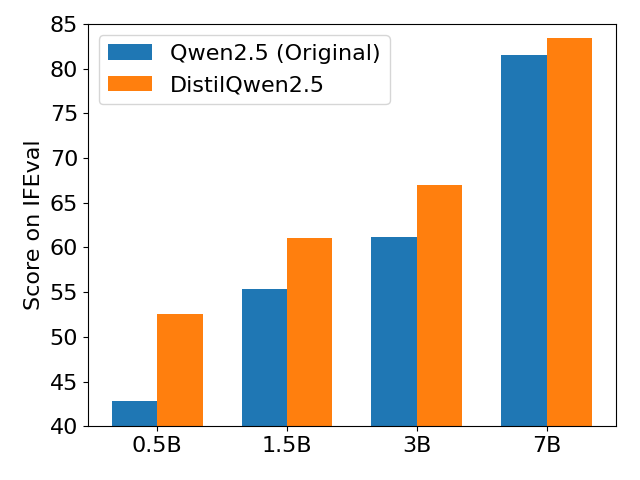}
        %\caption{IFEval}
    \end{subfigure}
    \caption{Brief comparison between original \emph{Qwen2.5} and \emph{DistilQwen2.5} models in terms of AlpacaEval 2.0 (length-controlled) and IFEval scores.}
    \label{fig:overall}
\end{figure}

In this paper, we introduce \emph{DistilQwen2.5}, a series of distilled LLMs derived from the \emph{Qwen2.5} models\footnote{\url{https://qwenlm.github.io/blog/qwen2.5/}}. In the beginning of the KD process, proprietary teacher LLMs, serving as multiple agents, are utilized to select, rewrite, and refine instruction-response pairs, tailoring them to be more conducive to learning by smaller student models. In particular, a Chain-of-Thought (CoT)~\cite{DBLP:conf/nips/Wei0SBIXCLZ22} rewriting approach is employed to significantly enhance the reasoning abilities of the distilled models. Beyond standard fine-tuning, we further introduce a model fusion approach to enable student models to incrementally integrate fine-grained hidden knowledge from their teacher models in a computationally efficient manner. This approach enhances the depth of understanding in student models beyond what black-box distillation processes can achieve.

In our experiments, we demonstrate that the resulting \emph{DistilQwen2.5} models show remarkable improvements in instruction-following performance across various NLP tasks compared to their original counterparts. Briefly, we present the AlpacaEval 2.0 (length-controlled)~\cite{DBLP:journals/corr/abs-2404-04475} and IFEval~\cite{DBLP:journals/corr/abs-2311-07911} scores of the \emph{DistilQwen2.5} models in Figure~\ref{fig:overall}. To enhance the public accessibility of our work, all models have been made available to the  open-source community. Furthermore, we describe several use cases to demonstrate the applications of our work in real-world scenarios.

\section{Related Work and Discussion}

Knowledge distillation (KD), originally proposed by \citet{DBLP:journals/corr/HintonVD15}, has emerged as a key technique for improving the efficiency of neural networks. Prior to the era of LLMs, several studies successfully demonstrated the distillation of BERT-based models~\cite{DBLP:journals/corr/abs-1910-01108,DBLP:conf/emnlp/JiaoYSJCL0L20,DBLP:conf/acl/SunYSLYZ20,DBLP:conf/acl/Pan0QZLH20,DBLP:conf/icassp/HouWCQFH23}, primarily focusing on specific NLP tasks. However, distillation for LLMs presents unique challenges due to the intricate dependencies among prediction tokens. In the literature, $f$-Distill~\cite{DBLP:conf/acl/Wen0DM23} minimizes a generalized $f$-divergence function for sequence-level KD. MiniLLM~\cite{DBLP:conf/iclr/Gu0WH24} introduces a reverse Kullback-Leibler divergence (KLD) objective to distill knowledge from white-box LLMs to student models. \citet{DBLP:conf/coling/WuTWY0W25} propose an adaptive approach that allocates weights to combine forward and reverse KLD objectives. FuseLLM~\cite{DBLP:conf/iclr/WanH0QB024} merges multiple powerful LLMs into a more capable student model.
%by integrating their diverse strengths.

Given that many powerful LLMs are accessible only through APIs, KD from proprietary LLMs to smaller open-source models (referred to as black-box KD) has garnered significant attention~\cite{DBLP:conf/acl/HsiehLYNFRKLP23}. To facilitate distillation from more advanced LLMs, some researchers leverage these models for data augmentation to fine-tune student LLMs~\cite{DBLP:journals/corr/abs-2412-04871}. \citet{DBLP:conf/acl/LiCCHGZ24} utilize the data selection capabilities of student LLMs to refine instruction-tuning data. \citet{DBLP:conf/iclr/Lou0XSAX0024} generate multi-faceted instructions for diverse tasks to enhance black-box KD. Additionally, \citet{DBLP:conf/emnlp/YueWHW24} propose a task-aware curriculum planning framework to improve instruction refinement.

In contrast to prior work, our approach emphasizes industrial practices that leverage the strengths of both black-box and white-box KD methods. Moreover, efficiency remains a critical barrier in industry, particularly for white-box KD. To address this, our work incorporates an efficient algorithm to integrate hidden knowledge from teacher models.

\begin{figure}
\centering
\includegraphics[width=.5\textwidth]{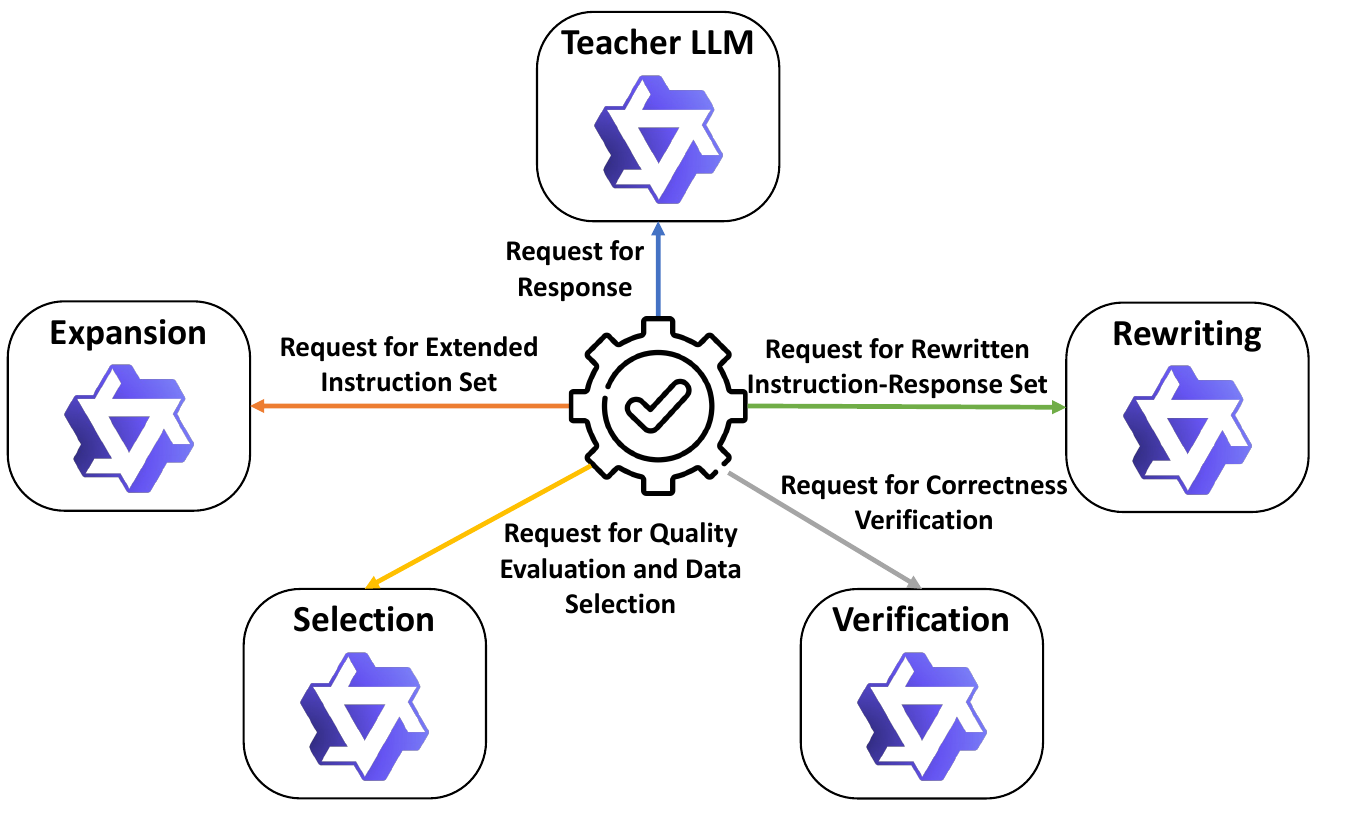}
\caption{Functionalities for LLMs/agents used in data augmentation and black-box distillation. \textbf{Disclaimer:} We use the Qwen logo in the figure; however, any LLMs with sufficient capabilities can be used as well.}
\label{fig:mt}
\end{figure}

\section{Our Approach}

In this section, we describe the industrial practices for distilling the \emph{DistilQwen2.5} models. 

\subsection{Multi-Agent Data Augmentation as Black-Box Knowledge Distillation}

We first leverage multi-agent data augmentation as black-box KD, where proprietary teacher models serve as the sources of knowledge. This approach is more computationally efficient than white-box KD and allows us to select more powerful proprietary models as teachers. In our work, we employ \emph{Qwen-max}\footnote{\url{https://qwenlm.github.io/}} to process the Chinese texts due to its strong capabilities in handling the Chinese language, and GPT-4/GPT-4o for other languages. In Figure~\ref{fig:mt}, we can see that a controller coordinates the entire pipeline of generating responses directly from the teacher model and invoking LLM agents to augment the training data. The functionalities of these LLM agents are described below.

\noindent\textbf{Expansion Agent.} The expansion agent is employed generate a diverse set of instruction variations, ensuring that student models are exposed to a comprehensive range of instructions. Importantly, it preserves the original NLP task category of the input instruction to prevent hallucinations and semantic drift caused by LLMs. For example, given the input ``Provide a brief overview of Newton's First Law of Motion'', the output could be ``Explain the meaning of Kepler's Third Law'', but not ``Give me a brief introduction to Albert Einstein’s life''. After instruction expansion, we also call the teacher model to generate responses for new instructions.

\noindent\textbf{Rewriting Agent.} The rewriting agent further enhances the quality and diversity of the training data. Unlike the expansion agent, the rewriting agent operates under stringent constraints to preserve the semantic integrity of the tasks expressed in instructions, ensuring that the rewritten content remains faithful to the original intent and task category. For example, the instruction ``Provide a summary of the economic impacts of climate change'' might be rewritten as ``Explain how climate change affects the economy''. Regarding the generated responses, we encourage them to be Chain-of-Thought (CoT) outputs for complex tasks such as logical reasoning, mathematical problems, and code generation \cite{DBLP:conf/nips/Wei0SBIXCLZ22}, as this significantly enhances the cognitive reasoning abilities of distilled, small models \cite{DBLP:conf/acl/HsiehLYNFRKLP23, DBLP:conf/emnlp/YueWHW24}.

\noindent\textbf{Selection Agent.} The selection agent automatically evaluates and chooses instruction-response pairs that are highly valuable for training the student model. This selection process is guided by various heuristic criteria, including informativeness, helpfulness, and potential for generalization to similar tasks. Additionally, we consider task balance when selecting these pairs, following the approach of \citet{DBLP:conf/emnlp/YueWHW24}. This guides the controller to filter out less useful data instances.

\noindent\textbf{Verification Agent.} Different from the selection agent, the verification agent is invoked each time new instruction-response instances are generated by LLMs to check the factual correctness.

Overall, the augmented dataset leverages a black-box KD method by encapsulating the distilled knowledge from larger models into training examples for student models. The distillation training process follows a supervised learning paradigm, utilizing the augmented instruction-response pairs.

\begin{table*}
\centering
\begin{scriptsize}
\begin{tabular}{lccccc}
\hline
\textbf{Model} & \textbf{AlpacaEval 2.0 (LC)} & \textbf{MT-Bench} & \textbf{MT-Bench (Single)}  & \textbf{IFEval (instruct-loose)} & \textbf{IFEval (strict-prompt)}\\
\hline
Qwen2.5-0.5B-Instruct & 2.46 & 5.49 & 6.26 & 42.81 & 30.31\\
\bf DistilQwen2.5-0.5B-Instruct$^*$ & 4.72 & 5.71 & 6.74 & 51.44 & 37.15\\
\bf DistilQwen2.5-0.5B-Instruct & \bf 4.89 & \bf 5.78 & \bf 6.83 & \bf 52.61 & \bf 37.82\\
\hline
Qwen2.5-1.5B-Instruct & 6.69 & 7.09 & 7.66 & 55.40 & 40.11\\
\bf DistilQwen2.5-1.5B-Instruct$^*$ & 13.30 & 7.27 & 7.90 & 60.63 & 73.02\\
\bf DistilQwen2.5-1.5B-Instruct & \bf 13.69 & \bf 7.35 & \bf 7.99 & \bf 61.10 & \bf 74.49\\
\hline
Qwen2.5-3B-Instruct & 17.98 & 7.92 & 8.40 & 61.18 & 74.58\\
\bf DistilQwen2.5-3B-Instruct$^*$ & 20.81 & 8.33 & 8.94 & 65.80 & 77.10\\
\bf DistilQwen2.5-3B-Instruct & \bf  20.91 & \bf  8.37 & \bf  8.97 & \bf  67.03 & \bf  77.36\\
\hline
Qwen2.5-7B-Instruct & 31.43 & 8.52 & 8.83 & 81.53 & 72.10\\
\bf DistilQwen2.5-7B-Instruct$^*$ & 34.78 & 8.75 & 9.19 & 83.41 & 73.20\\
\bf DistilQwen2.5-7B-Instruct & \bf 34.86 & \bf 8.76 & \bf 9.22 & \bf 83.48 & \bf 73.27\\
\hline
\end{tabular}
\end{scriptsize}
\caption{Performance comparison between the original \emph{Qwen2.5} model and the \emph{DistilQwen2.5} models in terms of instruction-following abilities across four parameter sizes: 0.5B, 1.5B, 3B, and 7B. Note: $^*$ indicates a variant of our model utilizing black-box KD over processed datasets. LC: length control (AlpacaEval 2.0).}
\label{tab:results}
\end{table*}

\subsection{Efficient Model Fusion as White-Box Knowledge Distillation}

In contrast to black-box KD, white-box KD involves having the student model mimic the distribution of the teacher model's logits, providing richer knowledge compared to learning from only the token with the highest output probability. In our work, we conduct white-box KD after the completion of black-box KD to maximize the utility of computational resources and aim to further improve the performance of student models by learning richer knowledge.
We assume that the student model, with learnable parameters $\theta$, has a probability function $p_S^\theta$ that is differentiable with respect to $\theta$. The token-level logits difference between $p_T$ (from the teacher model) and $p_S^\theta$ (from the student model) is defined as follows:

\begin{small}
\begin{equation}
D_{\theta}(x,y)=\frac{1}{L}\sum_{n=1}^{L} D_{\theta}\left(p_T(\cdot \mid y_{<n}, x) \parallel p_S^\theta(\cdot \mid y_{<n}, x)\right),
\end{equation}
\end{small}
where $x$ and $y$ denote the input and output sequences, respectively, and $L$ is the sequence length. The function $D_{\theta}(\cdot)$ can be any divergence measurement, such as KLD~\cite{DBLP:conf/iclr/Gu0WH24}, reverse KLD~\cite{DBLP:conf/coling/WuTWY0W25}, etc. The KD loss aims to minimize the divergence between the token sequences of the student and the teacher:
\begin{equation}
L(\theta)=\mathbb{E}_{(x,y) \sim (X,Y)}\left[D_{\theta}(x,y)\right].
\end{equation}

For industrial-scale implementation, it is infeasible to leverage existing white-box KD approaches such as those by \citet{DBLP:conf/iclr/Gu0WH24} and \citet{DBLP:conf/coling/WuTWY0W25}. The reasons are twofold: 
i) If the forward pass of the teacher model is performed simultaneously with the training of the student model, the GPU memory consumption becomes excessively high, especially when the teacher model is very large (e.g., 32B/72B).
ii) The vocabulary of the teacher and student models may not match, leading to a mismatch of the logits tensors of both models.

In our work, we observe that the sum of the probabilities of the top-10 tokens is almost equal to 1. This indicates that nearly all the knowledge of the teacher model is contained within the top-10 tokens. Therefore, we build a scalable white-box KD system that supports the following features:
i) A \emph{token alignment} operation \cite{DBLP:conf/iclr/WanH0QB024} is first conducted if the logits tensors of both models do not match.
ii) A distributed computing process is executed offline to generate the teacher model's logits with top-$K$ probabilities, where $K=10$ is set as default and adjustable for customized scenarios.
iii) A variant of $D_{\theta}(\cdot)$ is implemented where only the top-$K$ elements are calculated for divergence minimization.
Let $\mathbf{z}_{T}=[z_{T}^{(1)},z_{T}^{(2)},\cdots,z_{T}^{(K)}]$ and $\mathbf{z}_{S}=[z_{S}^{(1)},z_{S}^{(2)},\cdots,z_{S}^{(K)}]$ be the the top-$K$ logits from the teacher model, and the corresponding logits from the student model with matched indices in the vocabulary. The probabilities for computing $D_{\theta}(\cdot)$ is then calculated as follows:
$\mathbf{p}_T=\frac{\exp(\mathbf{z}_{T}/\mathcal{T})}{\sum_{k=1}^K \exp(z_{T}^{(k)}/\mathcal{T})}$,
$\mathbf{p}_S=\frac{\exp(\mathbf{z}_{S}/\mathcal{T})}{\sum_{k=1}^K \exp(z_{S}^{(k)}/\mathcal{T})}$,
where $\mathcal{T}$ is the temperature hyperparameter.
This approach not only reduces computation time but also improves the speed of storing and reading the logits, alleviating the storage pressure of our cloud computing system.

\section{Experimental Evaluation}

In this section, we present experimental setups and evaluation results of the \emph{DistilQwen2.5} models. 

\subsection{Experimental Setup}

The initial dataset consists of instruction-response pairs collected from several popular public datasets, including OpenHermes 2.5\footnote{\url{https://huggingface.co/datasets/teknium/OpenHermes-2.5}}, the Cleaned Alpaca Dataset\footnote{\url{https://github.com/gururise/AlpacaDataCleaned}}, and LCCD~\cite{DBLP:conf/nlpcc/WangKZHJZH20}, together with our in-house datasets. The pre-processing steps follow the method presented in~\cite{DBLP:journals/corr/abs-2412-04871}. Subsequently, the instruction-response pairs are carefully expanded, rewritten, verified and selected.
To create a series of smaller student LLMs, we utilize the~\emph{Qwen2.5} series as our backbone models, including their instruct versions with varying sizes: 0.5B, 1.5B, 3B, and 7B. The white-box teacher models are selected from \texttt{Qwen2.5-14B/32B/72B-Instruct}.
For student model distillation, the default learning rate and the epochs are set to $1 \times 10^{-5}$ and 3, respectively. We train all the models on a server equipped with eight A800 GPUs, each with 80GB memory.

\subsection{Evaluation Benchmarks}
AlpacaEval 2.0 (length-controlled)~\cite{DBLP:journals/corr/abs-2404-04475} assesses the instruction-following capabilities of LLMs across various domains. MT-Bench~\cite{DBLP:conf/acl/BaiLBHLZLSG0O24} is utilized to evaluate the multitasking abilities of our models. This benchmark challenges models with diverse tasks that require an understanding of multiple domains and the ability to quickly adapt to changing instructions, under both single-turn and multi-turn conversation settings. IFEval~\cite{DBLP:journals/corr/abs-2311-07911} assesses how models perform during dynamic user interactions. For rigorous comparison, we report the results in both instruct-loose and strict-prompt settings.

\subsection{Main Experimental Results}
The results of our experiments are summarized in Table~\ref{tab:results}. As illustrated, the \emph{DistilQwen2.5} models demonstrate superior performance across all benchmarks, outperforming both the baseline and original models by significant margins. Moreover, the proposed model fusion technique enhances the models' capabilities after the black-box KD process. We further observe that the improvement is more pronounced for smaller student backbones. Specifically, the improvement of \texttt{DistilQwen2.5-0.5B-Instruct} compared to \texttt{Qwen2.5-0.5B-Instruct} is larger than that of \texttt{DistilQwen2.5-7B-Instruct} compared to \texttt{Qwen2.5-7B-Instruct}.
This shows that the potential of smaller students is larger in terms using KD.
Overall, the experimental results empirically validate our distillation framework, demonstrating its effectiveness in enhancing the task-solving performance of lightweight LLMs.

\subsection{Detailed Analysis}

\subsection{Inference Speed of Teacher Logits Generation}
\begin{figure}
\centering
\includegraphics[width=.425\textwidth]{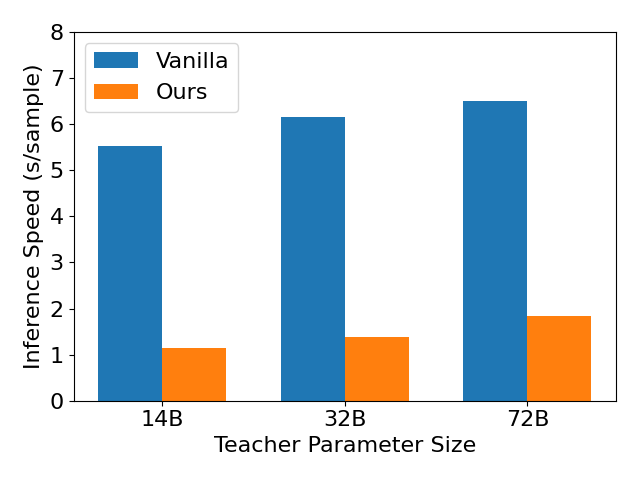}
\caption{Comparison of the inference speed for logits generation between our approach and the vanilla approach (average seconds per sample).}
\label{fig:infer}
\end{figure}

In our experiments, we measure the latency associated with generating logits across different sizes of teacher models, as shown in Figure~\ref{fig:infer}. Our implementation achieves a significantly accelerated inference speed, obtaining a 3$\times$ to 5$\times$ speedup compared to the vanilla implementation. Additionally, the reduction in logits does not lead to any noticeable decrease in the instruction-following abilities of the distilled smaller models, as revealed by our exploratory experiments.

\subsection{Fine-grained Model Capacity Analysis}
In this section, we provide a detailed capacity analysis of the \emph{DistilQwen2.5} models, leveraging the MT-bench benchmark~\cite{DBLP:conf/acl/BaiLBHLZLSG0O24} to quantify their performance across a diverse array of NLP tasks.
Due to space limitations, we show the results for two smallest models, with other models exhibiting similar trends. These results are detailed in Table~\ref{tab:mt}. Our analysis not only showcases the broad applicability of our \emph{DistilQwen2.5} models but also proves their enhanced capabilities and performance improvements over the original models.

\begin{table}
\centering
\begin{scriptsize}
\begin{tabular}{l | cc | cc}
\hline
\textbf{Task Type} & \textbf{0.5B} & \textbf{0.5B (Ours)} & \textbf{1.5B} & \textbf{1.5B (Ours)}\\
\hline
Writing & 6.08 & \bf 6.68 & \bf 8.38 & \bf 8.38\\
Roleplay & 7.07 & \bf 7.43 & 7.26 & \bf 8.13\\
Reasoning & 4 & \bf 4.2 & 3.9 & \bf 4.8\\
Mathematics & \bf 4.65 & \bf 4.65 & 6.85 & \bf 6.98\\
Coding & 4 & \bf 4.08 & 4.6 & \bf 5.04\\
Extraction & 3.55 & \bf 4.5 & 6.4 & \bf 6.6\\
STEM & 6.55 & \bf 6.83 & \bf 9.65 & 9.28\\
Humanity & \bf 8.1 & 7.95 & 9.73 & \bf 9.83\\
\hline
\end{tabular}
\end{scriptsize}
\caption{Detailed task-specific score comparisons between the original~\emph{Qwen2.5} and~\emph{DistilQwen2.5} models (0.5B and 1.5B) on MT-bench.}
\label{tab:mt}
\end{table}

\subsection{Comparison Against Other Small Models}
\begin{figure}
\centering
\includegraphics[width=.475\textwidth]{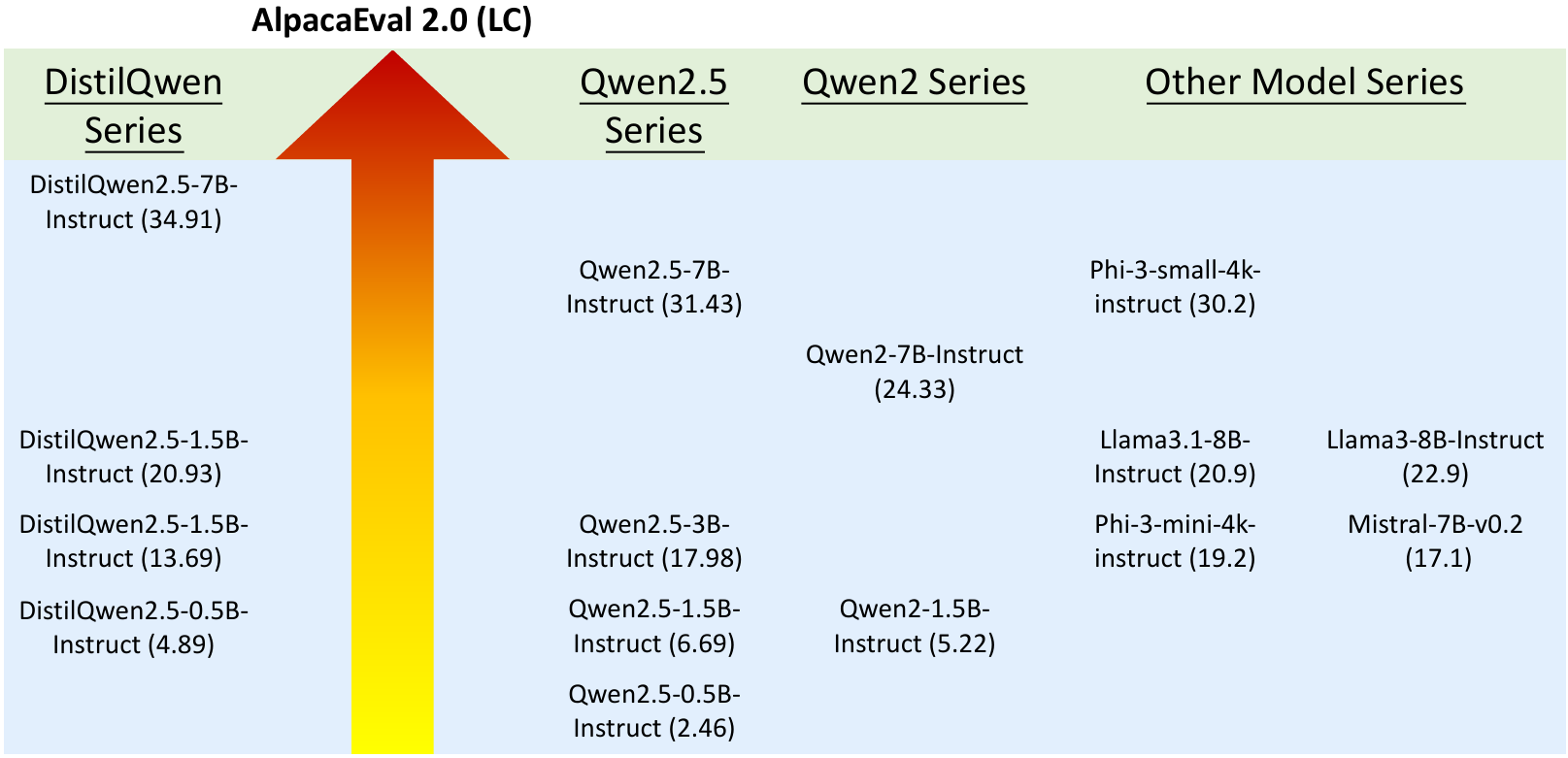}
\caption{Comparison between various small models (<10B) based on AlpacaEval 2.0 (length-controlled).}
\label{fig:alpca}
\end{figure}

To compare the performance against other models, we present the ranking in Figure~\ref{fig:alpca}. Notably, the~\emph{DistilQwen2.5} series demonstrates remarkable cost-effectiveness, achieving performance that closely rivals models with parameter sizes either approaching or exceeding double its own.

\begin{figure}
    \centering
    \begin{subfigure}{0.235\textwidth}
        \centering
        \includegraphics[width=\textwidth]{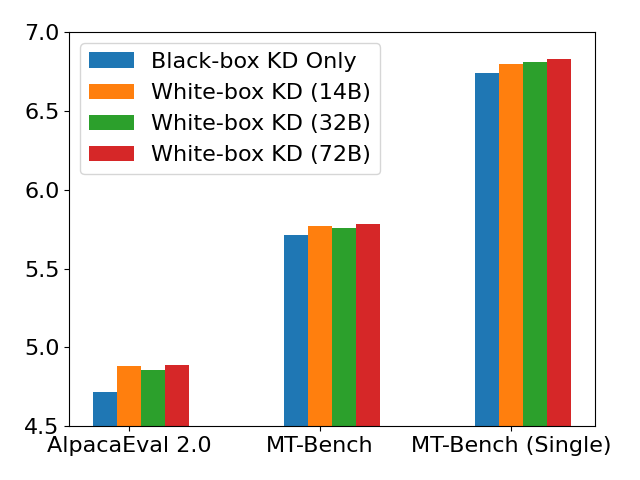}
        \caption{Student size: 0.5B}
    \end{subfigure}
    \hfill
    \begin{subfigure}{0.235\textwidth}
        \centering
        \includegraphics[width=\textwidth]{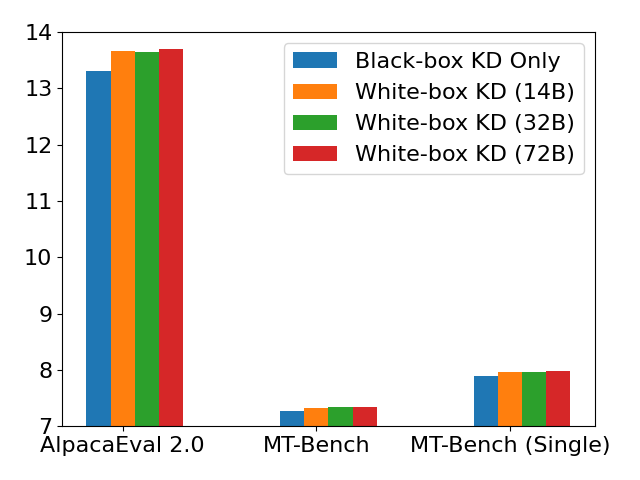}
        \caption{Student size: 1.5B}
    \end{subfigure}
    \caption{Comparison between black-box KD and white-box KD with varying teach model sizes after black-box KD, in terms of AlpacaEval 2.0 (length-controlled) and MT-Bench scores (both full and single).}
    \label{fig:mf}
\end{figure}

\begin{figure}
    \centering
    \begin{subfigure}{0.235\textwidth}
        \centering
        \includegraphics[width=\textwidth]{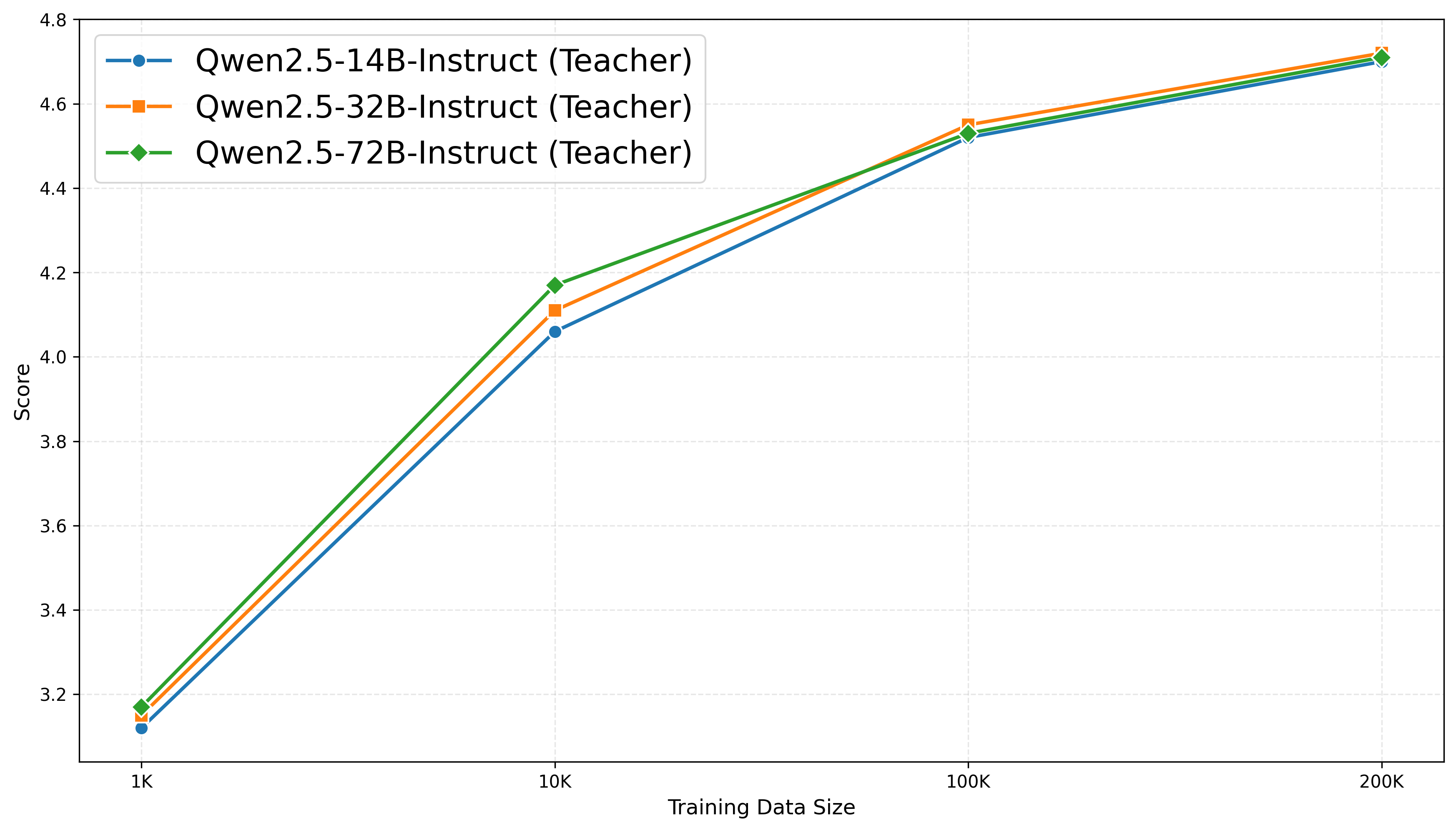}
        \caption{Student size: 0.5B}
    \end{subfigure}
    \hfill
    \begin{subfigure}{0.235\textwidth}
        \centering
        \includegraphics[width=\textwidth]{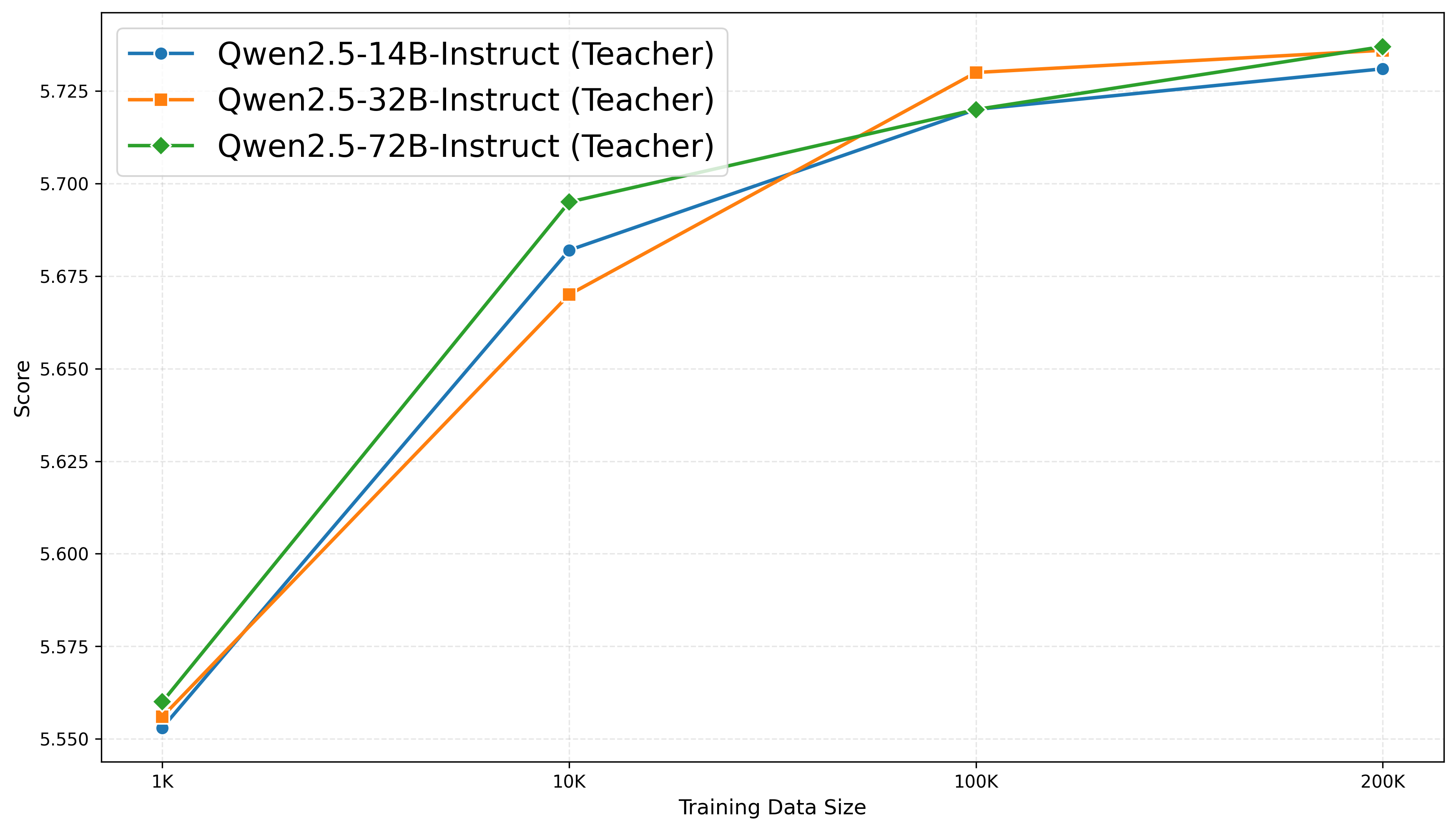}
        \caption{Student size: 0.5B}
    \end{subfigure}
    \begin{subfigure}{0.235\textwidth}
        \centering
        \includegraphics[width=\textwidth]{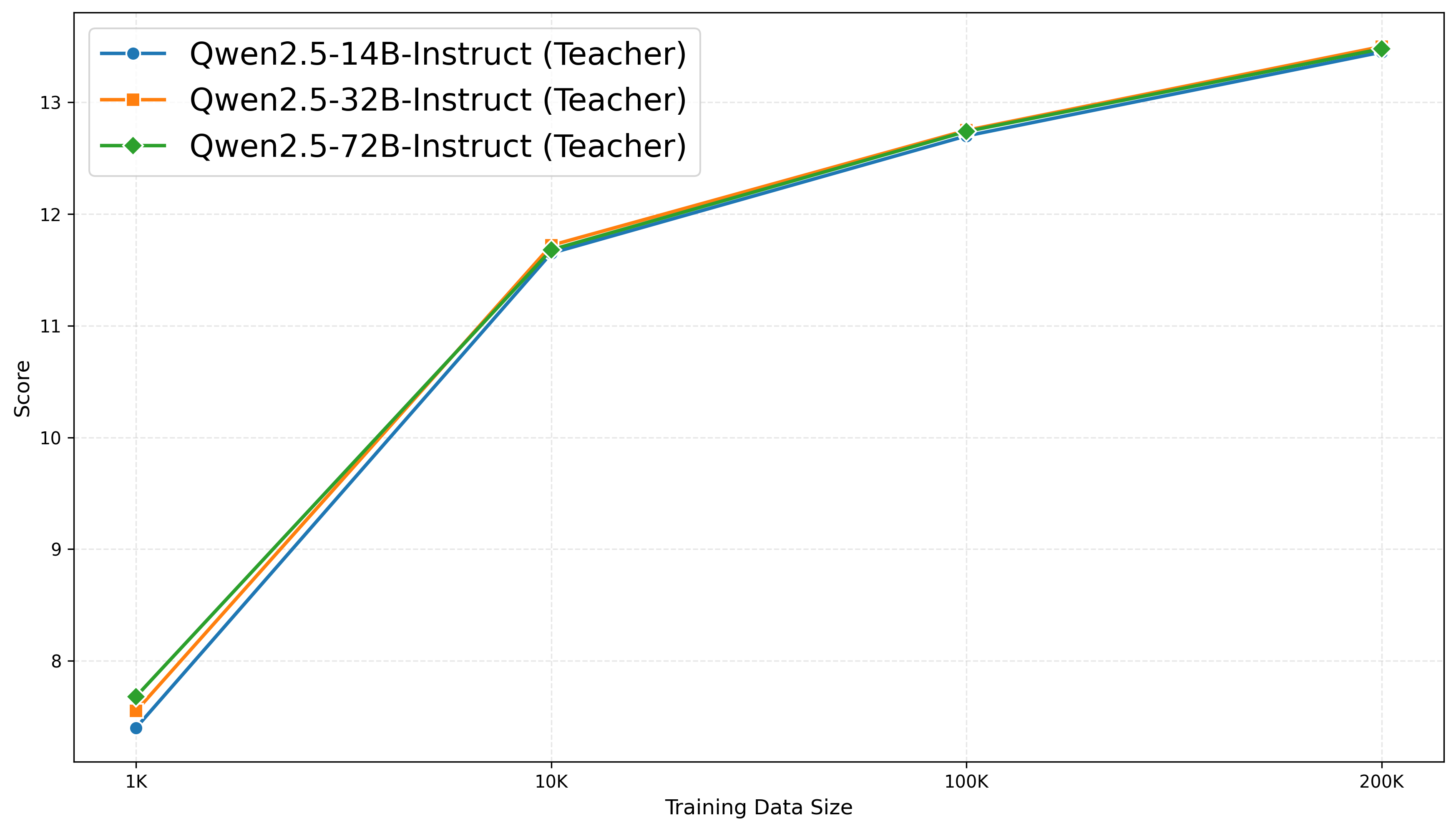}
        \caption{Student size: 1.5B}
    \end{subfigure}
    \hfill
    \begin{subfigure}{0.235\textwidth}
        \centering
        \includegraphics[width=\textwidth]{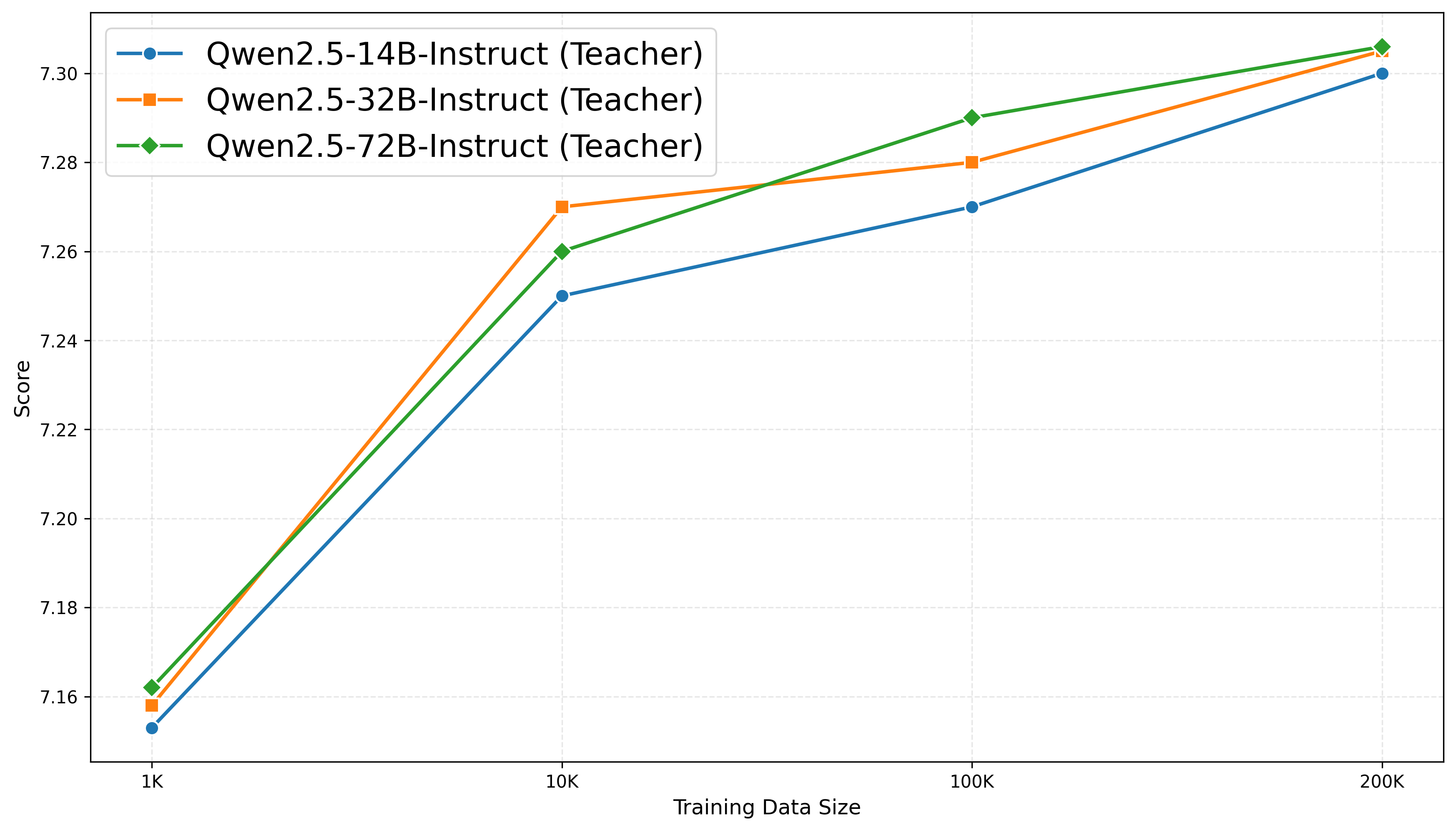}
        \caption{Student size: 1.5B}
    \end{subfigure}
    \caption{Performance of white-box KD with varying teach/student model sizes and dataset sizes.}
    \label{fig:mf2}
\end{figure}

\subsection{Analyzing the Parameter Sizes of Teacher LLMs for Model Fusion}
We conduct the first set of experiments following the completion of black-box KD. The results, presented in Figure~\ref{fig:mf}, demonstrate a trend of diminishing returns as teacher sizes increase (from 14B to 72B), indicating that larger teacher models offer limited improvements to the student model. This finding suggests that teacher models should not be excessively large to minimize computational costs.
The second set of experiments is conducted on model checkpoints without black-box KD, with results shown in Figure~\ref{fig:mf2}. We observe that as the dataset size increases, the improvement also gradually diminishes, indicating a diminishing return on additional data. 
However, notable improvements are observed with larger teacher models when the dataset comprises between 10K to 100K samples, suggesting that it can be more beneficial within the specific range.

\subsection{Additional Results}
Due to space limitations, we present several case studies in the appendix.

\section{Industrial Use Cases}

In addition to the~\emph{DistilQwen2.5} models presented, we outline two industrial use cases that illustrate the practical utility of our framework and models.

\subsection{SQL Completion for Big Data Platform}

In addition to instruction following, our framework can also address other tasks, such as code completion, which is also an auto-regressive task for LLMs. Here, we present a real-world application w.r.t. SQL completion. It helps users to formulate complex queries, optimize SQL statements, add conditions, or join tables based on existing queries. This technique significantly improves both the efficiency and accuracy of query composition and is widely utilized in our online big data platforms.

In the context of SQL completion for our big data platform, the primary evaluation metrics are \emph{Latency}, \emph{Pass@1} and \emph{Adoption Rate}. \emph{Latency} measures the system's speed in generating real-time suggestions as users input queries, whereas \emph{Pass@1} and \emph{Adoption Rate} reflect the utility and accuracy of the model's output based on automatic evaluation and human feedback. A key challenge is the trade-off between model scale and the performance metrics: although larger models can achieve higher adoption rates, they often result in increased inference time, which adversely affects latency. Therefore, the central optimization challenge for SQL completion in big data platforms lies in enhancing completion efficacy while maintaining a relatively compact model size.

During the initial deployment phase, we utilize the fine-tuned~\texttt{Qwen2.5-7B} model for deployment, which is quantized to~\texttt{int4} precision. By applying KD on a fixed dataset (i.e., an in-house SQL corpus), we obtain a~\texttt{Qwen2.5-3B} model. This model achieves a significant improvement, closely matching the performance of the 7B model, while increasing the inference speed by 1.4×. 
The online performance of these models is shown in Table~\ref{tab:sql}, where \emph{Adoption Rate} is obtained through online A/B testing on the big data platform.
Hence, our KD technique effectively balances performance and computational efficiency.

\subsection{KD Functionalities on AI Platform}

It should be acknowledged that our \emph{DistilQwen2.5} models are primarily designed for general domains. For domain-specific applications, further enhancement is necessary (as in the SQL completion case). To enable business users or LLM developers to distill their own models, we have integrated the continual KD feature together with the \emph{DistilQwen2.5} models into a cloud-native AI platform.

\begin{figure}
\centering
\includegraphics[width=.475\textwidth]{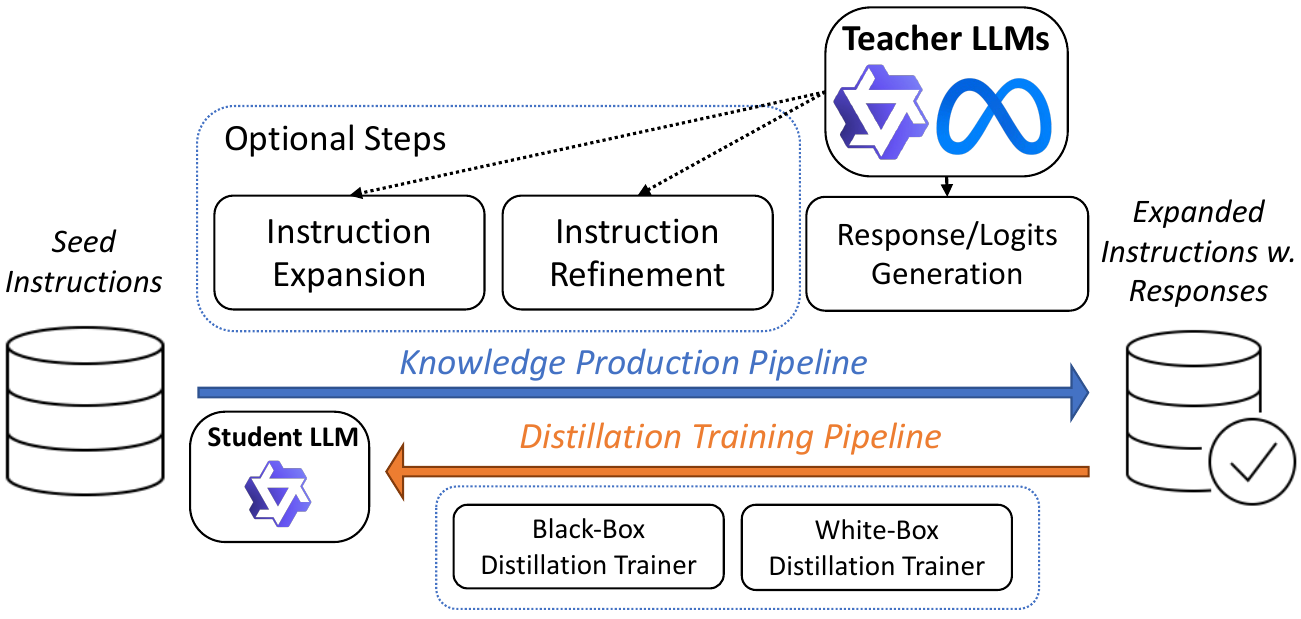}
\caption{Illustration of continual KD pipelines on the AI platform for business users or LLM developers.}
\label{fig:use}
\end{figure}

To facilitate seamless model optimization and customization, our AI platform provides robust KD functionalities, as shown in Fig.~\ref{fig:use}. It allows users to iteratively refine and tailor the \emph{DistilQwen2.5} models to specific domains. Key pipelines include: (1) the Knowledge Production Pipeline (KPP) and (2) the Distillation Training Pipeline (DTP). In KPP, optimal steps of instruction expansion and refinement can be applied to user-provided seed instructions from arbitrary domains. The teacher LLMs are then leveraged to generate responses or output logits according to user settings. In DTP, users can define custom training settings for either black-box or white-box distillation trainers, leveraging cloud resources for scalable distillation training. After that, the student model can be utilized for evaluation and deployment.

\begin{table}
\centering
\begin{small}
\begin{tabular}{l | ccc}
\hline
\textbf{Model Size} & \textbf{Latency} & \textbf{Pass@1} & \textbf{Adoption Rate}\\
& (ms) & & (\%)\\
\hline
7B (teacher) & 384 & 18.8 & 26.5\\
3B (student) & 148 & 17.9 & 25.5\\
\hline
\end{tabular}
\end{small}
\caption{Performance evaluation for SQL completion.}
\label{tab:sql}
\end{table}

\section{Conclusion and Future Work}
In this paper, we introduce \emph{DistilQwen2.5}, a family of distilled lightweight LLMs derived from the \emph{Qwen2.5} models. By leveraging both black-box and white-box KD techniques and efficient implementations and multiple agents, we demonstrate substantial improvements in model performance and real-world applications. For future work, we plan to investigate more diverse domain-specific applications to extend the practical impact of our framework. We also aspire to enhance the collaborative aspects of model fusion to allow for more dynamic knowledge transfer.

\section*{Limitations}

While the \emph{DistilQwen2.5} models demonstrate significant enhancements, several limitations remain that warrant further investigation. The distillation process hinges on the quality of the teacher models. Biases or errors inherent in the teacher models could propagate into the student models, potentially affecting their performance and fairness in specific contexts. Additionally, while we showcase domain-specific applications, the generalizability of our framework across diverse domains and languages remains to be thoroughly evaluated, which is beyond the scope of this work. Addressing these limitations will contribute to more robust LLMs tailored to a wider array of applications.

\section*{Ethical Considerations}

Distillation techniques make it feasible to deploy LLMs in resource-constrained environments, they also introduce the potential for bias and misinformation inherited from the teacher models. 
Additionally, the open-sourcing of \emph{DistilQwen2.5} models facilitates accessibility, but also raises concerns regarding misuse. Responsible use of the models requires establishing guidelines to prevent applications that may cause harm, violate privacy, or amplify malicious behavior.

%\section*{Acknowledgements}


\appendix

\section{Case Studies}
\label{sec:appendix}

In the following, we present several cases on the output comparison between original \emph{Qwen2.5} and \emph{DistilQwen2.5} models. In default, we leverage the 7B models for comparison.

\begin{table*}
\centering
\begin{small}
\begin{tabular}{l|l}
\toprule
\bf Question & Problem: \\
& If the system of equations: 6x-4y=a, 6y-9x=b. has a solution $(x, y)$ where $x$ and $y$ are both nonzero, find $\frac{{a}}{{b}}$, \\
& assuming $b$ is nonzero. \\
& Solution: \\
\midrule
\bf Ground Truth & -$\frac{{2}}{{3}}$ \\
\midrule
\bf Qwen2.5 & To find the value of a/b, we start by solving the system of equations for x and y in terms of a and b. \\ &
Given the system of equations 6x - 4y = a 6y - 9x = b, \\ &
First we can simplify these equations, Equation 1 can be written as 3x - 2y = a/2 \\ &
Equation 2 can be written as -3x + 6y = b,\\ &
Next we add equations 3 and 4 to eliminate x, (3x - 2y) + (-3x + 6y) = a/2 + b \\ &
which simplifies to 4y = a/2 + b. Solving for y, we get y = (a + 2b)/8. \\ &
Next we subtract equation 4 from equation 3 to eliminate y, \\ &
(3x - 2y) - (-3x + 6y) = a/2 - b, which simplifies to 6x - 8y = a/2 - b \\ &Substituting the value of y from equation 5 into this equation, we get 6x - 8((a + 2b)/8) = a/2 - b \\ &
which simplifies to 6x - (a + 2b) = a/2 - b or 6x = a + 2b + a/2 - b,\\ &
which further simplifies to 6x = (2a + 4b + a - 2b)/2 = (3a + 2b)/2. \\ &
Solving for x we get x = (3a + 2b)/12 = (a + (2b)/3)/4, \\ &
Now we need to find the ratio a/b From equation 5. We have y = (a + 2b)/8 From equation 6 we have\\ &
x = (3a + 2b)/12 \\ &
We substitute these expressions into the original equations. \\ &
Using equation 1 6x - 4y = a we substitute x = (3a + 2b)/12 \\ &
and y = (a + 2b)/8 6((3a + 2b)/12) - 4((a + 2b)/8) = a, which simplifies to \\ &(6(3a + 2b))/12 - (4(a + 2b))/8 = a \\ &
or (3a + 2b)/2 - (a + 2b)/2 = a, which further simplifies to (3a + 2b - a - 2b)/2 = a or (2a)/2 = a, \\ &
which is true This confirms our expressions are correct. From the simplified equations\\ &
we derive the ratio a/b = 2. \\ &
Thus the value of a/b is 2. \\
\midrule
\bf DistilQwen2.5 & To solve the given system of equations and find $\frac{{a}}{{b}}$, we start with the equations: \\
& 6x - 4y = a, \\
& 6y - 9x = b. \\
& First, we express x and y in terms of a and b. We can rewrite the equations as: \\
& 6x - 4y = a, \\
&-9x + 6y = b. \\
& To eliminate one of the variables, we can multiply the first equation by 3 and the second equation by 2: \\
&18x - 12y = 3a, \\
& -18x + 12y = 2b. \\
&Now, we add these two equations together: \\
& (18x - 12y) + (-18x + 12y) = 3a + 2b. \\
& This simplifies to: 0 = 3a + 2b. \\
& From this, we can solve for $\frac{{a}}{{b}}$ = -$\frac{{2}}{{3}}$ \\
\bottomrule
\end{tabular}
\end{small}
\caption{Output comparison between original \emph{Qwen2.5} and \emph{DistilQwen2.5} models.}
\label{tab:case1}
\end{table*}

\begin{table*}
\centering
\begin{small}
\begin{tabular}{l|l}
\toprule
\bf Question & What breed dog is smallest? \\
\midrule
\bf Qwen2.5 & The smallest breed of dog is generally considered to be the Chihuahua. \\ &
However, there are other very small breeds that are often mentioned, such as the Yorkipoo \\ &
(a mix between a Yorkshire Terrier and a Poodle),\\ &
Morkie (a mix between a Maltese and a Yorkshire Terrier), and Puggle (a mix between a Pug and a Beagle). \\ &
But in terms of pure breeds, the Chihuahua is the smallest. \\ &
Chihuahuas typically weigh between 2 and 6 pounds (0.9 to 2.7 kg) \\ &
and stand about 6 to 9 inches (15 to 23 cm) tall at the shoulder. \\
\midrule
\bf DistilQwen2.5 & The Chihuahua is generally recognized as the smallest dog breed in the world. \\
\bottomrule
\end{tabular}
\end{small}
\caption{Output comparison between original \emph{Qwen2.5} and \emph{DistilQwen2.5} models.}
\label{tab:case2}
\end{table*}

\begin{table*}
\centering
\begin{small}
\begin{tabular}{l|l}
\toprule
\bf Question & Write a product description for a new line of dog toys, called \"the squeaker\". \\ &
It's made of latex, and is designed for dogs of all ages and species. \\ &
It's also available in a variety of colors and shapes. \\ &
The response must contain at least 3 placeholders represented by square brackets, \\ &
such as [address], [name], and [phone number]. \\
\midrule
\bf Qwen2.5 & Introducing the [address] Pet Supply's newest addition to our toy line: the Squeaker!\\ &
Designed with love and play in mind, our Squeakers are crafted from high-quality, \\ &
durable latex that stands up to even the most enthusiastic chewers. \\ &
Whether your [name] is a playful puppy or a seasoned senior, our Squeakers are perfect for dogs\\ &
of all ages and species, providing endless hours of fun and stimulation. Available in a vibrant array of colors\\ &
and shapes, our Squeakers are not just toys; they're companions for playtime. \\ &
From the classic [shape1] to the whimsical [shape2], each design is carefully chosen to engage your [name]'s \\ &
curiosity and keep them entertained. \\ &
The unique squeaker mechanism ensures that every play session is filled with excitement and joy.\\ &
Order your [name]'s new favorite toy today by visiting our website or giving us a call at [phone number]. \\ &
We're here to help you find the perfect Squeaker to keep your [name] happy and healthy. Happy playing! \\
\midrule
\bf DistilQwen2.5 & Introducing The Squeaker – the ultimate playtime companion for your furry friend! \\ &
Crafted from durable, high-quality latex, \\ &
The Squeaker is designed to provide endless entertainment for dogs of all ages and breeds. \\ &
Available in a vibrant array of colors and fun shapes, \\ &
this toy is perfect for keeping your pet engaged and happy.\\ &
Each Squeaker features a built-in squeak mechanism that will capture your dog's attention\\ &
and stimulate their natural instincts to play and explore. \\ &
Whether your dog loves to chew, fetch, or simply cuddle with their toys, The Squeaker is sure to\\ &
become their new favorite.\\ &
Order yours today and watch your dog's tail wag with joy! \\ &
For more information, visit our website at [website], or contact our customer service team at [phone number]. \\ &
Don't miss out on the fun – make playtime special with The Squeaker![PetJoy] – Bringing joy to pets and\\ &
their owners. \\
\bottomrule
\end{tabular}
\end{small}
\caption{Output comparison between original \emph{Qwen2.5} and \emph{DistilQwen2.5} models.}
\label{tab:case3}
\end{table*}

\end{document}